\documentclass{article}
\pdfoutput=1

\usepackage[preprint]{corl_2026} 

\usepackage{amsmath}
\usepackage{amssymb}
\usepackage{booktabs}
\usepackage{graphicx}
\usepackage{subfig}
\usepackage{xcolor}
\usepackage{wrapfig}
\usepackage{multirow}
\usepackage{xspace}
\usepackage{float}
\usepackage{pifont}
\usepackage{tikz}
\usetikzlibrary{positioning, arrows.meta, shapes.geometric, fit, backgrounds, calc}
\usepackage[ruled,vlined]{algorithm2e}
\hypersetup{pdftitle={Beyond Imitation: Self-Improving Robot Policies via Off-Policy Q-Planning},
            pdfauthor={Varun Giridhar, Anant Khandelwal, Jeremy A. Collins, Ignat Georgiev, Animesh Garg},
            pdfsubject={}}

\newcommand{\qplan}{Q-Planning\xspace}
\newcommand{\cmark}{\textcolor[rgb]{0,0.5,0}{\ding{51}}}
\newcommand{\xmark}{\textcolor[rgb]{0.8,0.15,0.15}{\ding{55}}}

\newcommand{\R}{\mathbb{R}}

\newcommand{\va}{\mathbf{a}}
\newcommand{\vo}{\mathbf{o}}
\newcommand{\vz}{\mathbf{z}}
\newcommand{\vell}{\boldsymbol{\ell}}

\title{Beyond Imitation: Self-Improving Robot Policies via Off-Policy Q-Planning}

\author{%
  Varun Giridhar\quad Anant Khandelwal\quad Jeremy A.~Collins\\
  \bfseries Ignat Georgiev\quad Animesh Garg\\
  \mdseries Georgia Institute of Technology\\
  \texttt{\href{https://varungiridhar.github.io/qplanning/}{varungiridhar.github.io/qplanning}}
}

\begin{document}
\maketitle


\begin{abstract}
Behaviour Cloning (BC) has driven remarkable progress in robot manipulation, yet it is fundamentally limited by its inability to self-improve: a policy that fails cannot learn from that failure without additional human demonstrations. Reinforcement Learning fine-tuning offers a path to self-improvement but has proven difficult to scale to the multi-billion-parameter models underpinning modern robot policies. We propose \qplan{}, which equips a large visuomotor BC policy with a small off-policy Q-function. Because a Q-function estimates value rather than imitates actions, it can be trained on the same successful demonstrations as the BC policy and later absorb both successful and failed deployment rollouts, an asymmetry BC does not have. We exploit this asymmetry to enable value-guided action selection at inference (a single-step Q-weighted average over BC draws) and online self-improvement that fine-tunes only the Q-function, leaving the BC weights untouched. On LIBERO and bimanual RoboTwin, ten iterations of self-improvement lift every benchmark score we tested (LIBERO-10 $93\to99\%$, RoboTwin $83.8\to91.4\%$) and shorten successful episodes on the near-ceiling suites (LIBERO-Object, LIBERO-Goal). On two contact-rich bimanual real-robot tasks, the same loop (BC frozen, no human intervention) improves purely from its own deployment rollouts: stack-cups $40\to90\%$ and insert-wallet $25\to80\%$ in five iterations, whereas SFT on successful rollouts alone stalls at $55\%$ and $30\%$. Under an identical online budget \qplan{} is the only method, among Best-of-$N$, filtered SFT, IBRL, DSRL, and DAWR, that improves stably from failures without training an auxiliary actor.
\end{abstract}



\begin{figure}[H]
    \centering
    \includegraphics[width=0.95\textwidth]{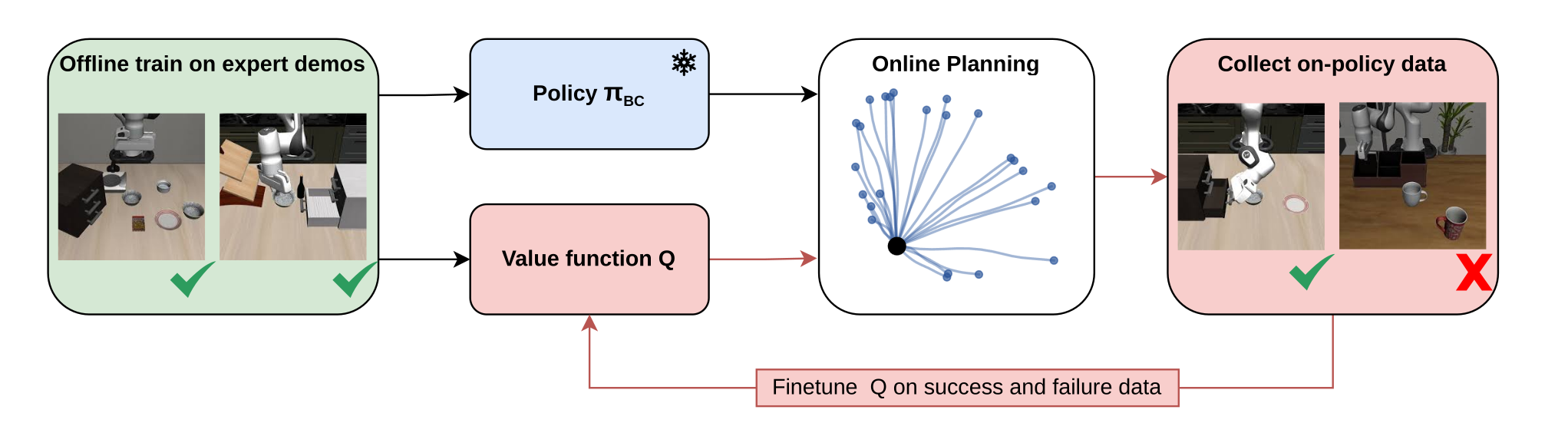}
    \caption{\textbf{\qplan{} overview.} \emph{Left:} at inference, the frozen BC policy samples $N$ candidate action chunks; the Q-function scores them and the executed chunk is a single-step Q-weighted average. \emph{Right:} rollouts (successful and failed) are appended to a replay buffer and used to fine-tune \emph{only} the Q-function. The updated $Q_\phi$ returns to the next inference iteration; the BC policy is never updated.}
    \label{fig:teaser}
\end{figure}

\section{Introduction}
\label{sec:introduction}


Robot manipulation has progressed rapidly through large-scale Behaviour Cloning (BC), with large visuomotor policies such as RT-2~\citep{brohan2023rt}, Octo~\citep{team2023octo}, $\pi^*_{0.6}$~\citep{intelligence2025pi}, Diffusion Policy~\citep{chi2023diffusion} and FastWAM~\citep{yuan2026fast} demonstrating impressive zero-shot generalisation on real-world tasks. This success, however, comes at a structural cost: BC policies are bounded by the data they were trained on. A trained BC policy that fails on a task cannot learn from that failure without additional human teleoperation, and there is no mechanism by which rollouts collected during deployment improve subsequent behaviour. As we scale toward models with billions of parameters trained on millions of demonstrations, this \emph{demo ceiling} becomes the dominant bottleneck for further progress.

The dominant response has been Reinforcement Learning fine-tuning of the policy itself, pursued aggressively in the past year~\citep{chen2025pirl,li2025simplevla,lu2025vla,chen2025conrft,li2025gr,guo2025improving,tan2025interactive,mark2024policy,liu2026can}. Yet updating a multi-billion-parameter vision-language-action (VLA) model on sparse-reward, on-policy data is expensive, brittle, and can silently degrade the very BC prior that gave the policy its competence in the first place~\citep{kumar2021should}. An alternative line of work, value-based planning~\citep{williams2017model,hansen2022temporal,hansen2023td,schrittwieser2020mastering}, sidesteps policy updates by re-ranking action candidates at inference time, but has not been demonstrated at VLA scale, where high-dimensional flow-matching action spaces and long horizons make planning over random proposals intractable.

\textbf{Our insight} is that the actor and the critic can be decoupled \emph{and trained on different data}. A BC policy can only be trained on successful demonstrations (failure trajectories are not what we want to imitate), but an off-policy Q-function has no such restriction: it can be trained on any trajectory, successful or failed, because it estimates value rather than imitates actions. This asymmetry is what makes the self-improvement loop efficient: we can keep the expensive BC policy frozen and absorb all deployment signal, including failures, into a much smaller off-policy Q-function. Figure~\ref{fig:teaser} illustrates this loop.

To combine the two components at inference, we draw $N$ candidate action chunks from the frozen BC policy and execute a single-step Q-weighted average, with softmax weights taken over the Q-scores of the candidates. The Q-function's vision and language encoders are amortised once per planning step, so only its action-conditioned decoder scales with $N$. Together, an off-policy Q-function and a Q-weighted average over BC draws turn a frozen BC policy into a self-improving system. Ten iterations of online self-improvement lift LIBERO-10 success from $93\%$ to $\mathbf{99\%}$ and bimanual RoboTwin from $83.8\%$ to $\mathbf{91.4\%}$; on a real bimanual robot the same loop improves purely from deployment rollouts on two contact-rich tasks, stack-cups $40\to90\%$ and insert-wallet $25\to80\%$, over five iterations. Our contributions are:
\begin{enumerate}
    \item \textbf{An off-policy Q-function for large BC policies.} A Q-chunking architecture with HL-Gauss categorical outputs and its own DinoV2~\citep{oquab2023dinov2} and T5~\citep{raffel2020exploring} encoders, trained on the same demonstrations used to train the BC policy.
    \item \textbf{A real-time value-guided action selector.} A single-step Q-weighted average over $N$ BC flow-matching draws, amortising the encoders across all candidates so a planning step costs $400$ ms on RoboTwin ($1.6\times$ faster than a single $10$-step BC inference and comfortably inside the $960$ ms replan budget; App.~\ref{app:latency}).
    \item \textbf{Self-improvement from failures without touching the BC.} A loop that folds successful \emph{and failed} deployment rollouts into Q-only updates, evaluated across every LIBERO suite, bimanual RoboTwin, and two contact-rich real-robot tasks.
\end{enumerate}


\section{Related work}
\label{sec:related-work}


\textbf{Large-scale BC and VLA policies.} The dominant paradigm for general-purpose robot manipulation is large-scale Behaviour Cloning on teleoperated demonstrations~\citep{brohan2022rt,brohan2023rt,team2023octo,padalkar2023open,chi2023diffusion,intelligence2025pi,yuan2026fast,bousmalis2023robocat}. These policies excel at imitating the demonstration distribution but inherit its ceiling: they cannot incorporate failure observations collected during deployment, and out-of-distribution states reliably degrade performance~\citep{kumar2021should,laskey2017dart}. \qplan{} treats such a BC policy as a frozen action proposal distribution and gives it a value function it never had.

\textbf{Value functions and planning for control.} Q-learning~\citep{mnih2013playing,haarnoja2018soft,lillicrap2015continuous} and value-based planners~\citep{williams2017model,hansen2022temporal,hansen2023td,schrittwieser2020mastering} have demonstrated strong control performance in low-dimensional or simulated regimes, but extending them to VLA-scale manipulation has been hindered by the cost of online search over high-dimensional action spaces. Recent work has explored pre-trained values as guidance for robotic foundation models~\citep{nakamoto2024steering,bhateja2023robotic,ma2022vip,ma2023liv,wagenmaker2025steering}, but does not address how to use the value function to actively self-improve via online iteration. \qplan{} closes this gap.

\textbf{Self-improvement and RL fine-tuning of robot policies.} An active body of work fine-tunes large robot policies with online~\citep{chen2025pirl,li2025simplevla,lu2025vla,chen2025conrft,guo2025improving,li2025gr,tan2025interactive,liu2026can,xiao2025self} or offline~\citep{kumar2022pre,mark2024policy,yang2024robot,hu2025rac} RL, and several recent systems iteratively co-improve a policy with a world model~\citep{guo2026vlaw,liu2026self,jia2026dreamplan,bousmalis2023robocat}. All of these methods update the policy parameters, often the entire VLA, which is expensive at scale and risks degrading the BC prior. \qplan{} departs from this consensus: only the Q-function is updated during self-improvement, while the BC policy is left untouched.

\textbf{Positioning against value-guidance and imitation-bootstrapped methods.} Table~\ref{tab:methods} distinguishes \qplan{} along five axes. V-GPS~\citep{nakamoto2024steering} steers a frozen policy with an \emph{offline}-trained value and never iterates online. DSRL~\citep{wagenmaker2025steering} adapts online but through a latent-noise auxiliary actor. IBRL~\citep{hu2023ibrl} keeps its IL policy frozen but trains a fresh RL actor
from scratch, which at VLA scale means re-learning behaviour the BC already
encodes; DAWR, the advantage-weighted regression baseline from the DPPO
paper~\citep{ren2024dppo}, fine-tunes the policy weights directly, which is
expensive at scale and risks degrading the BC prior. Concurrent BC-to-Q work~\citep{dodeja2026bcq} shares the frozen-BC + off-policy-Q ingredients but trains an auxiliary RL actor and has not been demonstrated on a multi-billion-parameter VLA. \qplan{} is the only entry that is simultaneously BC-frozen, plug-and-play with a multi-billion-parameter policy, self-improving from \emph{failure} rollouts, and free of an auxiliary actor.

\begin{table}[t]
\centering
\setlength{\tabcolsep}{6pt}
\renewcommand{\arraystretch}{1.25}
\caption{\textbf{Positioning against the closest methods.} \qplan{} is the only method that combines all five axes.}
\label{tab:methods}
\begin{tabular}{@{}lccccc@{}}
\toprule
Method & \begin{tabular}{@{}c@{}}BC\\frozen\end{tabular} & \begin{tabular}{@{}c@{}}Plug-and-play\\large VLA\end{tabular} & \begin{tabular}{@{}c@{}}Learn from\\failures\end{tabular} & \begin{tabular}{@{}c@{}}Offline-to-\\online $Q$\end{tabular} & \begin{tabular}{@{}c@{}}No aux.\\actor\end{tabular} \\
\midrule
V-GPS~\citep{nakamoto2024steering}   & \cmark & \cmark & \xmark & \xmark & \cmark \\
DSRL~\citep{wagenmaker2025steering}  & \cmark & \cmark & \cmark & \cmark & \xmark \\
IBRL~\citep{hu2023ibrl}              & \cmark & \xmark & \cmark & \xmark & \xmark \\DAWR~\citep{ren2024dppo}             & \xmark & \xmark & \cmark & \xmark & \cmark \\
\textbf{\qplan{} (ours)}             & \cmark & \cmark & \cmark & \cmark & \cmark \\
\bottomrule
\end{tabular}
\end{table}


\section{Method}
\label{sec:method}

\textbf{Overview.} \qplan{} consists of three components: an off-policy Q-function over action chunks (Sec.~\ref{sec:method-q}), a single-step Q-weighted average over BC action-chunk draws (Sec.~\ref{sec:method-plan}), and a self-improvement loop that fine-tunes only the Q-function (Sec.~\ref{sec:method-si}). At every stage the base BC policy is frozen.

\subsection{Problem setting}
\label{sec:method-setup}

We address language-conditioned manipulation in a partially-observed MDP with action $\va_t \in \R^{d_a}$, observation $\vo_t$ (RGB images from one or more cameras), natural-language instruction $\vell$, and per-step terminal indicator $d_t \in \{0,1\}$ together with sparse terminal reward $r_t \in \{0, 1\}$ that is $1$ only on the timestep at which the task is completed successfully. Following recent VLA work~\citep{chi2023diffusion,intelligence2025pi,yuan2026fast}, the BC policy $\pi^{\text{BC}}$ outputs an \emph{action chunk} $\va_{t:t+H} \in \R^{H \times d_a}$ of length $H$ at each planning step. We assume access to a dataset $\mathcal{D}_{\text{BC}}$ of successful demonstrations used to train $\pi^{\text{BC}}$. Our goal is to improve policy performance with minimal additional human supervision: we may collect rollouts with our own planner and update value-side parameters, but we never re-collect demonstrations and never update $\pi^{\text{BC}}$.

\subsection{Off-policy Q-function over action chunks}
\label{sec:method-q}

\textbf{Motivation.} Standard scalar Q-regression under sparse terminal rewards is unstable~\citep{farebrother2024stop}: most transitions carry no reward signal, and the long horizons of manipulation tasks amplify bootstrapping error. We therefore (i) treat the length-$H$ action chunk as a single super-action so the effective bootstrapping horizon shrinks by a factor of $H$ (Q-chunking~\citep{li2026reinforcement}), and (ii) replace scalar regression with HL-Gauss categorical regression~\citep{farebrother2024stop}, which projects each scalar target onto a fixed grid of bins via a Gaussian kernel and trains the Q-network with a cross-entropy objective.

\textbf{Module design.} The Q-function has its own visual and language encoders, parameter-disjoint from the BC policy: $\vz^{\text{vis}}_t = E^{\text{vis}}_\phi(\vo_t)$ uses DinoV2 and $\vz^{\text{txt}} = E^{\text{txt}}_\phi(\vell)$ uses T5. The Q-network $Q_\phi(\vo_t, \vell, \va_{t:t+H})$ is a transformer decoder that cross-attends to $[\vz^{\text{vis}}_t; \vz^{\text{txt}}]$ and takes the candidate action chunk as a sequence of query tokens. The decoder output is projected to $B$ HL-Gauss logits $\ell_\phi$; the predicted scalar value is the expectation of the softmaxed categorical distribution over the bin grid $\{v_b\}_{b=1}^B$ of bin centres uniformly spaced in $[v_{\min}, v_{\max}]$:
\begin{equation}
Q_\phi(\vo_t, \vell, \va_{t:t+H}) = \sum_{b=1}^{B} v_b \cdot \mathrm{softmax}\!\big(\ell_\phi(\vo_t, \vell, \va_{t:t+H})\big)_b .
\end{equation}

\textbf{Training.} Following the Q-chunking formulation~\citep{li2026reinforcement}, we train $Q_\phi$ on chunked transitions $(\vo_t, \va_{t:t+H}, r_{t:t+H}, \vo_{t+H}, \va_{t+H:t+2H})$ sampled from a buffer $\mathcal{D}$, initialised to the demonstration dataset $\mathcal{D}_{\text{BC}}$ and later expanded with online rollouts (Sec.~\ref{sec:method-si}). At training time we do not query $\pi^{\text{BC}}$ for a bootstrap action: we
use the next chunk $\va_{t+H:t+2H}$ directly from the buffer, and since a chunk
drawn from $\mathcal{D}_{\text{BC}}$ is exactly what $\pi^{\text{BC}}$ was
trained to imitate, the target remains an unbiased estimate of
$Q^{\pi^{\text{BC}}}$. The chunked squared-error Bellman loss is
\begin{equation}
\label{eq:q-target}
\mathcal{L}(\phi) = \mathbb{E}_{\mathcal{D}}\!\left[\bigg(Q_\phi(\vo_t, \vell, \va_{t:t+H}) - \sum_{t'=0}^{H-1}\gamma^{t'}\,r_{t+t'} - \gamma^{H}\,Q_{\bar\phi}(\vo_{t+H}, \vell, \va_{t+H:t+2H})\bigg)^{\!2}\right] ,
\end{equation}
where $\bar\phi$ are exponential-moving-average target parameters. In practice we do not minimise this squared error directly: following HL-Gauss \citep{farebrother2024stop}, we project the scalar target onto the bin grid and minimise the cross-entropy to the predicted bin distribution, which stabilises learning under sparse, bimodal returns. A failed rollout is simply an all-zero-reward trajectory whose terminal chunk
does not bootstrap, pinning its target to zero. Critically, this loss applies unchanged when $\mathcal{D}$ later includes online rollouts containing failures (Sec.~\ref{sec:method-si}), because $Q^{\pi^{\text{BC}}}$ is well-defined for any state-action input regardless of where the data came from. This is the asymmetry that lets us train $Q_\phi$ on data $\pi^{\text{BC}}$ itself could not be trained on. App.~\ref{app:bootstrap} justifies the choice of the one-step-shifted target over a planner-selected bootstrap.

Keeping the Q-function's encoders parameter-disjoint from $\pi^{\text{BC}}$ (DinoV2 and T5 backbones, App.~\ref{app:arch}) is what makes the self-improvement loop safe: failures in one network cannot corrupt the other, so the Q-function can be updated freely without ever touching the BC weights.

\subsection{Q-weighted action selection}
\label{sec:method-plan}

\textbf{Motivation.} Given $Q_\phi$, the natural inference-time strategy is to sample $N$ action chunks per planning step, score them with $Q_\phi$, and execute a value-informed aggregation. Two properties of the modern BC head make a very simple selector sufficient at VLA scale: flow-matching draws are multi-modal, so the $N$ candidates cover \emph{distinct} action modes rather than sitting inside a Gaussian envelope around a single mean; and they lie on the manifold the BC already finds plausible, so every candidate is inside $Q_\phi$'s training support. Multi-modality is important for self-improvement: any behaviour the BC head can produce with non-negligible probability is a candidate the planner can select and the loop can amplify, so exploration is not restricted to a single Gaussian ball around the BC mode. We therefore avoid iterative sample-based search entirely (e.g., MPPI; Sec.~\ref{sec:exp-planning}, App.~\ref{app:temporal}) and instead run a single-step Q-weighted average over the raw BC draws.

\textbf{Module design.} At each planning step $t$ we draw $N$ candidate chunks $\va^{(n)}_{t:t+H} \sim \pi^{\text{BC}}(\vo_t, \vell)$ from a truncated $3$-step flow-matching pass of the BC policy (the reduced denoising budget both lowers latency and preserves sample diversity relative to a full $10$-step draw), score each with $Q_\phi$, and execute a softmax Q-weighted average with temperature $\lambda$:
\begin{equation}
\label{eq:qweight}
w^{(n)} \propto \exp\!\big(Q_\phi(\vo_t, \vell, \va^{(n)}_{t:t+H}) / \lambda\big),
\qquad
\bar{\va}_{t:t+H} = \sum_{n=1}^N w^{(n)} \, \va^{(n)}_{t:t+H}.
\end{equation}
$\bar{\va}_{t:t+H}$ is executed in the environment up to a replanning interval. The Q-network is evaluated for all $N$ candidates in a single batched forward pass: vision and text tokens are encoded once per planning step (about $25$ ms combined) and the $N$ candidates enter the decoder as $N$ parallel query sequences, so only the $\sim$500M Q-decoder cost scales with $N$ (roughly $2$-$3$ ms per candidate). A full planning step, including BC draws, encoding, scoring, and aggregation, takes $400$ ms on the bimanual RoboTwin setup with $N{=}32$. This is $1.6\times$ faster than a single $10$-step BC inference and comfortably inside the $960$ ms replan budget, the time before the executed portion of the current chunk runs out; App.~\ref{app:latency} breaks down the full latency profile.


\subsection{Self-improvement loop}
\label{sec:method-si}

A BC policy trained only on successful demonstrations has never seen the failure modes that emerge under autonomous execution. \qplan{} closes this loop by deploying the planner, collecting both successful \emph{and} failed rollouts, and refining \emph{only} the Q-function on them (Algorithm~\ref{alg:si}). Each iteration runs the planner for $M$ episodes per task, appends the collected transitions to a replay buffer $\mathcal{D}$, and refines $Q_\phi$ for $S$ gradient steps on $\mathcal{D}$ via Eq.~\ref{eq:q-target}. Because only $Q_\phi$ ($\sim 1$B parameters, App.~\ref{app:impl}) is updated, one iteration is dramatically cheaper than a full-policy gradient step.

\begin{algorithm}[H]
\caption{\qplan{} Self-Improvement}
\label{alg:si}
\SetAlgoLined
$\mathcal{D} \leftarrow \mathcal{D}_{\text{BC}}$\;
\For{iteration $i = 1, 2, \ldots$}{
  \tcp*[h]{(1) Collect rollouts under \qplan{}}\;
  \For{task $k \in \mathcal{T}$, episode $j = 1 \ldots M$}{
    Reset env; receive $\vo_0$\;
    \While{episode not done}{
      Draw $N$ chunks $\{\va^{(n)}_{t:t+H}\}$ from $\pi^{\text{BC}}$ and compute $\bar{\va}_{t:t+H}$ via Eq.~\ref{eq:qweight}\;
      Execute $\bar{\va}_{t:t+H}$; observe $r_{t:t+H}, \vo_{t+H}$\;
    }
    Append the full episode to $\mathcal{D}$\;
  }
  \tcp*[h]{(2) Refine Q only (BC frozen)}\;
  \For{$s = 1 \ldots S$}{
    Update $\phi$ on a minibatch from $\mathcal{D}$ via $\mathcal{L}(\phi)$ (Eq.~\ref{eq:q-target})\;
    EMA update $\bar\phi \leftarrow \eta\, \phi + (1{-}\eta)\, \bar\phi$\;
  }
}
\end{algorithm}

We use $M{=}100$ episodes per task per iteration in simulation ($20$ per task on hardware) and $S{=}200$ Q-only gradient steps per iteration; full hyperparameters are in Appendix~\ref{app:impl}.



\section{Experimental results}
\label{sec:result}

Our experiments target four questions: \textbf{(Q1)} What action source should the Q-function score, i.e., does the choice of proposal distribution matter? \textbf{(Q2)} Does the online self-improvement loop lift performance? \textbf{(Q3)} Does \qplan{} \emph{enable} that self-improvement relative to value-guidance and imitation-bootstrapped alternatives? \textbf{(Q4)} Does self-improvement translate to the real world?

\subsection{Experimental setup}
\label{sec:exp-setup}

We evaluate on two simulated manipulation benchmarks: the four LIBERO~\citep{liu2023libero} suites (Spatial, Object, Goal, and the $10$ hard long-horizon tasks of LIBERO-$10$), and $47$ RoboTwin~\citep{mu2025robotwin} tasks (a subset of the full RoboTwin suite that we have working evaluations for). All reported success rates and episode lengths aggregate over $20$ episodes per task. For the BC policy we use the publicly-released FastWAM~\citep{yuan2026fast} weights, frozen throughout. The Q-function is trained on the same successful demonstrations that FastWAM was trained on (no additional data), for $12$k gradient steps on LIBERO and $45$k on RoboTwin, on $4 \times$ H200 GPUs, until per-task Q-values converge. LIBERO is evaluated with the standard episode budget of $540$ environment steps (rather than the $700$ steps used in the FastWAM paper); absolute LIBERO success rates therefore sit a few percentage points below FastWAM's reported numbers, and this protocol applies identically to both baseline and \qplan{} rows.

\subsection{Q1: which action source for planning?}
\label{sec:exp-planning}

We compare three candidates for the $N$ chunks $Q_\phi$ scores: \emph{MPPI}~\citep{williams2017model} with zero-mean Gaussian noise around the BC mean ($T{=}3$ iterations); \emph{MPPI + temporal smoothing}, the same variant with noise convolved by a $1$-D Gaussian kernel so perturbed chunks stay temporally coherent (App.~\ref{app:temporal}); and our single-step Q-weighted average over multi-modal $3$-step flow-matching draws (Sec.~\ref{sec:method-plan}). All share the same $Q_\phi$ and $N{=}64$; the unguided FastWAM policy is included as a reference. Evaluated on LIBERO-10.

\begin{table}[h]
\centering
\small
\setlength{\tabcolsep}{5pt}
\caption{\textbf{Action-source comparison on LIBERO-10} ($10$ tasks, $20$ episodes/task). Rows below the reference share the same $Q_\phi$ and differ only in how candidates are drawn. Latency is median wall-clock time per planning step on an NVIDIA L40S under LIBERO's eval-parity settings (App.~\ref{app:latency}). Temporally-smoothed MPPI and our single-step Q-weighted average match on success ($93.0\%$), but ours runs $1.7\times$ faster and has fewer hyper-parameters.}
\label{tab:planning}
\begin{tabular}{lccc}
\toprule
Action source & Success $\uparrow$ & Mean ep.\ length $\downarrow$ & Latency (ms) $\downarrow$ \\
\midrule
FastWAM (reference, no Q)                     & 90.0          & 274           & 646  \\
MPPI                                          & 89.5          & 270           & 1114 \\
MPPI + temporal smoothing                     & \textbf{93.0} & \textbf{261}  & 1114 \\
Ours (3-step FM draws + Q-weighted avg.)      & \textbf{93.0} & \textbf{261}           & \textbf{640}  \\
\bottomrule
\end{tabular}
\end{table}

Two observations. Vanilla MPPI sits \emph{below} the unguided BC baseline: jagged noise produces chunks outside $Q_\phi$'s training support, so the planner over-weights candidates with spuriously high Q-values. Temporally-smoothed MPPI and our single-step Q-weighted average over multi-modal flow-matching draws both recover the same $+3$pp gain over the frozen BC; we adopt the latter for the rest of the paper because it removes three hand-crafted hyper-parameters and runs $1.7\times$ faster on LIBERO and $3.2\times$ faster on RoboTwin (App.~\ref{app:latency}), which is what makes the self-improvement loop in Sec.~\ref{sec:exp-si} and the real-robot deployment in Sec.~\ref{sec:exp-real} real-time.

\subsection{Online self-improvement with Q-Planning}
\label{sec:exp-si}

Table~\ref{tab:offline-main} reports results across the four LIBERO suites and RoboTwin under three settings: the frozen FastWAM baseline, \qplan{} evaluated \emph{offline} (Q-weighted selection over BC draws, no self-improvement), and \qplan{} after $10$ iterations of online self-improvement (Q-only updates, BC frozen throughout). The offline column already shows the $+1.3$pp mean lift value guidance buys over the frozen BC ($4$ of $5$ benchmarks won), but the focus of this section is what happens \emph{after} the self-improvement loop is switched on.

\begin{table}[h]
\centering
\small
\setlength{\tabcolsep}{4pt}
\caption{\textbf{Main results} on the four LIBERO suites and RoboTwin. Per-row values are means of per-task success rates ($\uparrow$) and per-task mean successful episode length ($\downarrow$). \textbf{Bold} marks the best success per row. \qplan{} (offline) improves over the BC policy on $4$ of $5$ benchmarks by $+1.3$pp on average; $10$ iterations of Q-only self-improvement then lift every benchmark, reaching $97.6\%$ mean success ($+5.5$pp over the frozen BC).}
\label{tab:offline-main}
\begin{tabular}{lcccccc}
\toprule
 & \multicolumn{2}{c}{FastWAM~\citep{yuan2026fast}} & \multicolumn{2}{c}{\qplan{} (offline)} & \multicolumn{2}{c}{\qplan{} (self-improved)} \\
\cmidrule(lr){2-3} \cmidrule(lr){4-5} \cmidrule(lr){6-7}
Benchmark & Succ. & Len. & Succ. & Len. & Succ. & Len. \\
\midrule
LIBERO-Spatial       & 90.5           & 106 & 91.5           & 115 & \textbf{98.5}  & 107 \\
LIBERO-Object        & 100.0          & 138 & 99.5           & 139 & \textbf{100.0} & 120 \\
LIBERO-Goal          & 97.0           & 107 & 99.0           & 110 & \textbf{99.0}  &  99 \\
LIBERO-10            & 90.0           & 274 & 93.0           & 261 & \textbf{99.0}  & 224 \\
RoboTwin ($47$ tasks)& 83.2           & 220 & 83.8           & 231 & \textbf{91.4}  & 232 \\
\midrule
Mean                 & 92.1           & --  & 93.4           & --  & \textbf{97.6}  & --  \\
\bottomrule
\end{tabular}
\end{table}

\begin{figure}[t]
    \centering
    \includegraphics[width=\textwidth]{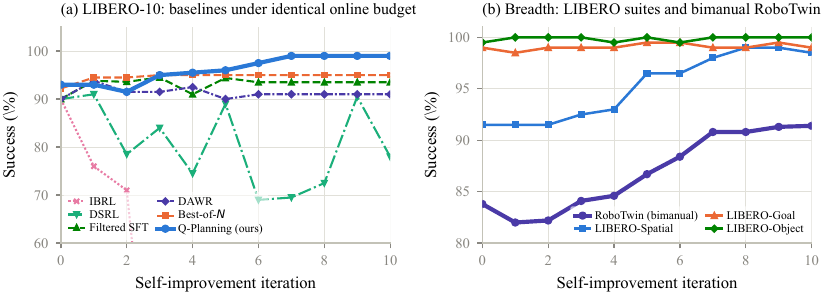}
    \caption{\textbf{\qplan{} self-improvement in simulation.} \emph{(a)} LIBERO-10 vs. five online baselines under an identical rollout budget: \qplan{} lifts success from $93\%$ to $\mathbf{99\%}$ while every baseline plateaus, oscillates, or collapses. \emph{(b)} Per-iteration trajectories across every LIBERO suite and bimanual RoboTwin, visualising the endpoints reported in Table~\ref{tab:offline-main}. The BC policy is frozen throughout; only $Q_\phi$ is updated.}
    \label{fig:si-sim}
\end{figure}

\textbf{Q2: The self-improvement loop lifts every benchmark.} With $M{=}100$ episodes per task per iteration and $S{=}200$ Q-only gradient steps per iteration (Algorithm~\ref{alg:si}), the loop takes every benchmark to a higher success rate: LIBERO-Spatial $91.5\to98.5\%$, LIBERO-10 $93\to99\%$, RoboTwin $83.8\to\mathbf{91.4}\%$, and mean $92.1\to\mathbf{97.6}\%$ ($+5.5$pp over the frozen BC and $+4.2$pp over the offline \qplan{} column). On the two suites that were already near ceiling under the offline planner (LIBERO-Object, LIBERO-Goal), success has no room to grow, so the loop instead \emph{shortens} successful episodes ($139\to120$ and $110\to99$ steps). Figure~\ref{fig:si-sim}(b) plots the per-iteration trajectory of these gains.

\begin{figure}[t]
    \centering
    \includegraphics[width=\textwidth]{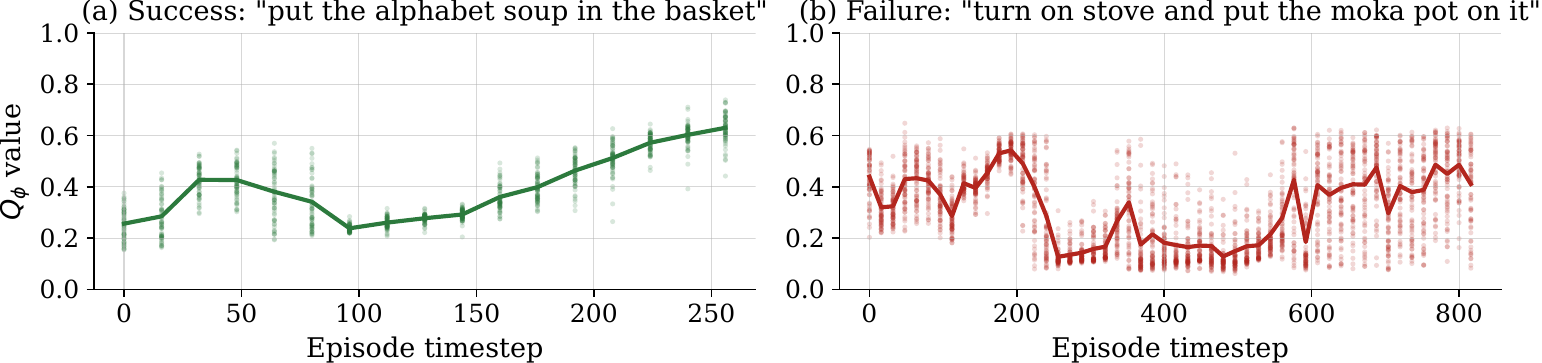}
    \caption{\textbf{Q-value over time} on two representative LIBERO-10 episodes. Solid lines show $Q_\phi$ of the executed (Q-weighted average) chunk; faint scatter shows the $N{=}64$ BC candidates scored per planning step.}
    \label{fig:q-over-time}
\end{figure}

\textbf{Q3: \qplan{} is what enables this self-improvement.} Figure~\ref{fig:si-sim}(a) compares \qplan{} on LIBERO-10 against five online baselines under the same rollout budget: \emph{Best-of-$N$} (same self-improvement loop, argmax selection instead of the weighted average), \emph{Filtered SFT} (re-imitating only successful rollouts), \emph{IBRL}~\citep{hu2023ibrl}, \emph{DSRL}~\citep{wagenmaker2025steering}, and \emph{DAWR}~\citep{ren2024dppo}. Best-of-$N$ plateaus at $95\%$, so value-guided \emph{selection} alone captures part of the gain but only the Q-weighted average plus self-improvement carries \qplan{} to $99\%$. Filtered SFT plateaus at $93.5\%$, confirming that re-imitating only successes cannot absorb the failure signal $Q_\phi$ can. IBRL collapses, DSRL swings widely between $69\%$ and $91\%$, and DAWR hovers below the frozen BC. Under the same online budget \qplan{} is the only method that improves stably from failures. Figure~\ref{fig:q-over-time} visualises $Q_\phi$'s predictions: on success it climbs monotonically from $\sim0.26$ to $\sim0.63$, on failure it oscillates in a lower range, and the candidate scatter shows it separating high- from low-value BC draws at every step, the signal both the Q-weighted average and the loop exploit.

\subsection{Real-robot self-improvement}
\label{sec:exp-real}

\begin{figure}[t]
    \centering
    \begin{minipage}[c]{0.43\textwidth}
        \centering
        \includegraphics[height=0.9in]{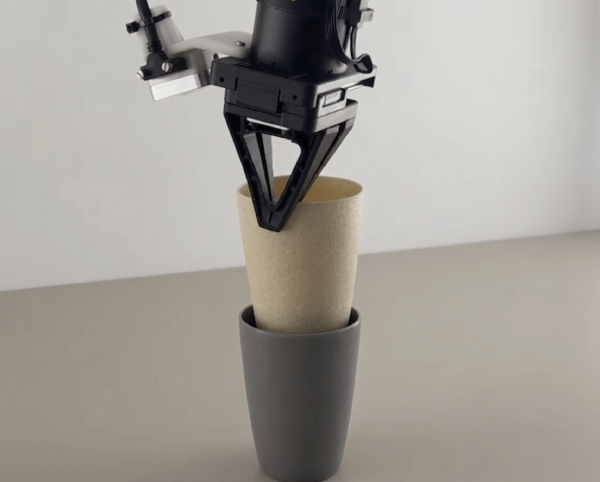}\hspace{2mm}\includegraphics[height=0.9in]{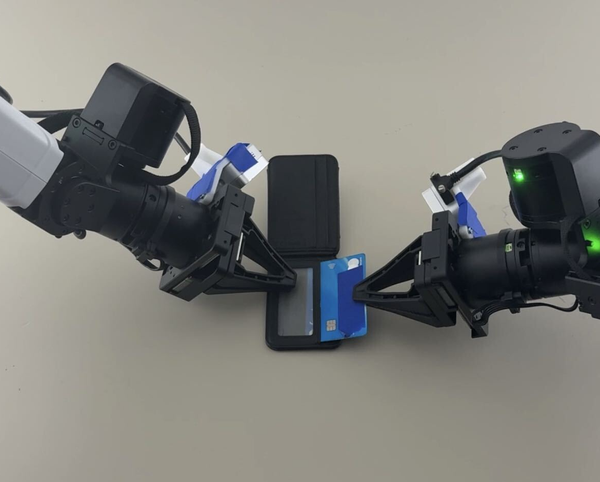}
        \\[2pt]
        {\footnotesize \emph{stack-cups}\hspace{1.0in}\emph{insert-wallet}}
    \end{minipage}
    \hfill
    \begin{minipage}[c]{0.55\textwidth}
        \centering
        \includegraphics[width=\linewidth]{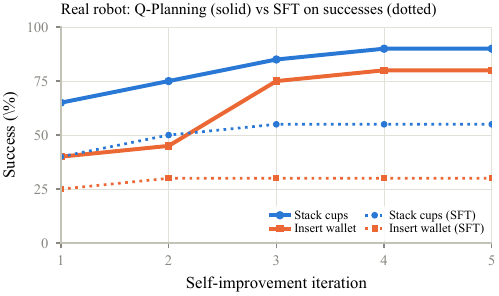}
    \end{minipage}
    \caption{\textbf{Real-robot self-improvement on two contact-rich bimanual tasks.} From $100$ base demos and $20$ online episodes per iteration, with the BC frozen and no human intervention, \qplan{} improves from its own failures: stack-cups $40\to 90\%$ and insert-wallet $25\to80\%$ in five iterations (Q-weighted selection alone starts the curves at $65\%$ and $40\%$). SFT on successes alone (dotted) stalls at $55\%$ and $30\%$.}
    \label{fig:si-real}
\end{figure}

\textbf{Q4: Self-improvement translates to the real world.} We deploy \qplan{} on two contact-rich bimanual tasks (Fig.~\ref{fig:si-real}; setup in App.~\ref{app:real}): \emph{stack-cups} (stacking plastic cups) and \emph{insert-wallet} (slotting a credit card into a wallet). From $100$ base demonstrations and $20$ online episodes per iteration for $5$ iterations, with the BC frozen and no human intervention, \qplan{} lifts stack-cups from its $40\%$ BC baseline to $90\%$ and from $25\%$ to $80\%$ on the harder insert-wallet task. SFT on successful rollouts alone (dotted in Fig.~\ref{fig:si-real}) stalls at $55\%$ and $30\%$: the gains come from the failure signal $Q_\phi$ absorbs and SFT discards.


\section{Conclusion}
\label{sec:conclusion}

We introduced \qplan{}, which makes a frozen BC policy self-improving by
exploiting a single asymmetry: BC must train on successful demonstrations,
while an off-policy Q-function can train on any rollout. In practice this
asymmetry buys a lot: a $\sim1$B-parameter Q-function trained beside a
multi-billion-parameter policy selects among BC draws in real time and absorbs
every deployment rollout into Q-only updates that never
touch the BC weights. Ten such iterations lift every simulated benchmark we
tested (mean success $92.1\to97.6\%$), shortening episodes where success was
already at ceiling; five lift two contact-rich real-robot tasks (stack-cups
$40\to90\%$, insert-wallet $25\to80\%$) with no human intervention. Because
only $Q_\phi$ is updated, self-improvement scales with the critic's size
rather than the policy's, an increasingly attractive property as VLA backbones
grow toward $10$B$+$ parameters. Nor is the framework intrinsically tied to
imitation learning: any proposal distribution that produces diverse plausible
action chunks (offline RL, reward-weighted imitation, world-model rollouts)
could stand in for the BC head, a natural next step for \qplan{}.

\section{Limitations}
\label{sec:limitations}

\textbf{BC policy dependence and exploration boundary.} \qplan{} cannot bootstrap from scratch: any behaviour the BC head cannot produce with non-negligible probability is a behaviour the planner cannot select and the loop cannot learn to exploit. Multi-modal flow-matching draws widen this envelope beyond a single Gaussian around the BC mode, but tasks where the base BC generates \emph{no} successful chunks at all remain out of reach. Gains therefore scale with BC quality and diversity.

\textbf{Q-decoder scaling with $N$.} The BC and Q encoders are amortised once per planning step, but the Q decoder scales linearly with the candidate count $N$ ($\sim 2$-$3$ ms per candidate on an L40S). Our deployed configurations ($N{=}64$ on LIBERO, $N{=}32$ on RoboTwin) fit inside the frozen BC baseline's latency budget (Table~\ref{tab:planning}, App.~\ref{app:latency}), but pushing $N$ toward the hundreds to attack harder exploration would eventually make the decoder pass, not the BC draw, the bottleneck.

\textbf{Terminal-reward supervision.} Our loop assumes a per-episode success detector: the environment success bit in simulation and a human-provided per-episode label on the real robot. Extending to open-ended tasks would require language-conditioned or learned success models.


\clearpage
\acknowledgments{We are grateful to the Georgia Tech PACE Phoenix cluster for compute, and to the FastWAM~\citep{yuan2026fast} authors for releasing pretrained weights and evaluation code that made this study possible.}


\bibliography{main}  

\begin{thebibliography}{47}
\providecommand{\natexlab}[1]{#1}
\providecommand{\url}[1]{\texttt{#1}}
\expandafter\ifx\csname urlstyle\endcsname\relax
  \providecommand{\doi}[1]{doi: #1}\else
  \providecommand{\doi}{doi: \begingroup \urlstyle{rm}\Url}\fi

\bibitem[Zitkovich et~al.(2023)Zitkovich, Yu, Xu, Xu, Xiao, Xia, Wu, Wohlhart,
  Welker, Wahid, Vuong, Vanhoucke, Tran, Soricut, Singh, Singh, Sermanet,
  Sanketi, Salazar, Ryoo, Reymann, Rao, Pertsch, Mordatch, Michalewski, Lu,
  Levine, Lee, Lee, Leal, Kuang, Kalashnikov, Julian, Joshi, Irpan, Ichter,
  Hsu, Herzog, Hausman, Gopalakrishnan, Fu, Florence, Finn, Dubey, Driess,
  Ding, Choromanski, Chen, Chebotar, Carbajal, Brown, Brohan, Arenas, and
  Han]{brohan2023rt}
B.~Zitkovich, T.~Yu, S.~Xu, P.~Xu, T.~Xiao, F.~Xia, J.~Wu, P.~Wohlhart,
  S.~Welker, A.~Wahid, Q.~Vuong, V.~Vanhoucke, H.~Tran, R.~Soricut, A.~Singh,
  J.~Singh, P.~Sermanet, P.~R. Sanketi, G.~Salazar, M.~S. Ryoo, K.~Reymann,
  K.~Rao, K.~Pertsch, I.~Mordatch, H.~Michalewski, Y.~Lu, S.~Levine, L.~Lee,
  T.-W.~E. Lee, I.~Leal, Y.~Kuang, D.~Kalashnikov, R.~Julian, N.~J. Joshi,
  A.~Irpan, B.~Ichter, J.~Hsu, A.~Herzog, K.~Hausman, K.~Gopalakrishnan, C.~Fu,
  P.~Florence, C.~Finn, K.~A. Dubey, D.~Driess, T.~Ding, K.~M. Choromanski,
  X.~Chen, Y.~Chebotar, J.~Carbajal, N.~Brown, A.~Brohan, M.~G. Arenas, and
  K.~Han.
\newblock Rt-2: Vision-language-action models transfer web knowledge to robotic
  control.
\newblock In J.~Tan, M.~Toussaint, and K.~Darvish, editors, \emph{Proceedings
  of The 7th Conference on Robot Learning}, volume 229 of \emph{Proceedings of
  Machine Learning Research}, pages 2165--2183. PMLR, 06--09 Nov 2023.
\newblock URL \url{https://proceedings.mlr.press/v229/zitkovich23a.html}.

\bibitem[{Octo Model Team} et~al.(2024){Octo Model Team}, Ghosh, Walke,
  Pertsch, Black, Mees, Dasari, Hejna, Xu, Luo, Kreiman, Tan, Chen, Sanketi,
  Vuong, Xiao, Sadigh, Finn, and Levine]{team2023octo}
{Octo Model Team}, D.~Ghosh, H.~Walke, K.~Pertsch, K.~Black, O.~Mees,
  S.~Dasari, J.~Hejna, C.~Xu, J.~Luo, T.~Kreiman, Y.~Tan, L.~Y. Chen,
  P.~Sanketi, Q.~Vuong, T.~Xiao, D.~Sadigh, C.~Finn, and S.~Levine.
\newblock Octo: An open-source generalist robot policy.
\newblock In \emph{Proceedings of Robotics: Science and Systems}, Delft,
  Netherlands, 2024.

\bibitem[Intelligence et~al.(2025)Intelligence, Amin, Aniceto, Balakrishna,
  Black, Conley, Connors, Darpinian, Dhabalia, DiCarlo,
  et~al.]{intelligence2025pi}
P.~Intelligence, A.~Amin, R.~Aniceto, A.~Balakrishna, K.~Black, K.~Conley,
  G.~Connors, J.~Darpinian, K.~Dhabalia, J.~DiCarlo, et~al.
\newblock {$\pi^{*}_{0.6}$}: a {VLA} that learns from experience.
\newblock \emph{arXiv preprint arXiv:2511.14759}, 2025.

\bibitem[Chi et~al.(2023)Chi, Feng, Du, Xu, Cousineau, Burchfiel, and
  Song]{chi2023diffusion}
C.~Chi, S.~Feng, Y.~Du, Z.~Xu, E.~Cousineau, B.~Burchfiel, and S.~Song.
\newblock Diffusion policy: Visuomotor policy learning via action diffusion.
\newblock \emph{arXiv preprint arXiv:2303.04137}, 2023.

\bibitem[Yuan et~al.(2026)Yuan, Dong, Liu, and Zhao]{yuan2026fast}
T.~Yuan, Z.~Dong, Y.~Liu, and H.~Zhao.
\newblock Fast-wam: Do world action models need test-time future imagination?
\newblock \emph{arXiv preprint arXiv:2603.16666}, 2026.

\bibitem[Chen et~al.(2025)Chen, Liu, Zhang, Guo, Xu, Lin, Zang, Zhang, Yu, Fan,
  et~al.]{chen2025pirl}
K.~Chen, Z.~Liu, T.~Zhang, Z.~Guo, S.~Xu, H.~Lin, H.~Zang, Q.~Zhang, Z.~Yu,
  G.~Fan, et~al.
\newblock $\pi$rl: Online rl fine-tuning for flow-based vision-language-action
  models.
\newblock \emph{arXiv preprint arXiv:2510.25889}, 2025.

\bibitem[Li et~al.(2025)Li, Zuo, Yu, Zhang, Yang, Zhang, Zhu, Zhang, Chen, Cui,
  et~al.]{li2025simplevla}
H.~Li, Y.~Zuo, J.~Yu, Y.~Zhang, Z.~Yang, K.~Zhang, X.~Zhu, Y.~Zhang, T.~Chen,
  G.~Cui, et~al.
\newblock Simplevla-rl: Scaling vla training via reinforcement learning.
\newblock \emph{arXiv preprint arXiv:2509.09674}, 2025.

\bibitem[Lu et~al.(2025)Lu, Guo, Zhang, Zhou, Jiang, Gao, Tang, and
  Wang]{lu2025vla}
G.~Lu, W.~Guo, C.~Zhang, Y.~Zhou, H.~Jiang, Z.~Gao, Y.~Tang, and Z.~Wang.
\newblock Vla-rl: Towards masterful and general robotic manipulation with
  scalable reinforcement learning.
\newblock \emph{arXiv preprint arXiv:2505.18719}, 2025.

\bibitem[Chen et~al.(2025)Chen, Tian, Liu, Zhou, Li, and Zhao]{chen2025conrft}
Y.~Chen, S.~Tian, S.~Liu, Y.~Zhou, H.~Li, and D.~Zhao.
\newblock Conrft: A reinforced fine-tuning method for vla models via
  consistency policy.
\newblock \emph{arXiv preprint arXiv:2502.05450}, 2025.

\bibitem[Li et~al.(2025)Li, Ma, Xu, Cui, Cui, Han, Huang, Kong, Liu, Niu,
  et~al.]{li2025gr}
Y.~Li, X.~Ma, J.~Xu, Y.~Cui, Z.~Cui, Z.~Han, L.~Huang, T.~Kong, Y.~Liu, H.~Niu,
  et~al.
\newblock Gr-rl: Going dexterous and precise for long-horizon robotic
  manipulation.
\newblock \emph{arXiv preprint arXiv:2512.01801}, 2025.

\bibitem[Guo et~al.(2025)Guo, Zhang, Chen, Ji, Wang, Hu, and
  Chen]{guo2025improving}
Y.~Guo, J.~Zhang, X.~Chen, X.~Ji, Y.-J. Wang, Y.~Hu, and J.~Chen.
\newblock Improving vision-language-action model with online reinforcement
  learning.
\newblock In \emph{2025 IEEE International Conference on Robotics and
  Automation (ICRA)}, pages 15665--15672. IEEE, 2025.

\bibitem[Tan et~al.(2025)Tan, Dou, Zhao, and
  Kr{\"a}henb{\"u}hl]{tan2025interactive}
S.~Tan, K.~Dou, Y.~Zhao, and P.~Kr{\"a}henb{\"u}hl.
\newblock Interactive post-training for vision-language-action models.
\newblock \emph{arXiv preprint arXiv:2505.17016}, 2025.

\bibitem[Mark et~al.(2024)Mark, Gao, Sampaio, Srirama, Sharma, Finn, and
  Kumar]{mark2024policy}
M.~S. Mark, T.~Gao, G.~G. Sampaio, M.~K. Srirama, A.~Sharma, C.~Finn, and
  A.~Kumar.
\newblock Policy agnostic rl: Offline rl and online rl fine-tuning of any class
  and backbone.
\newblock \emph{arXiv preprint arXiv:2412.06685}, 2024.

\bibitem[Liu et~al.(2026)Liu, Gao, Wei, Chen, Liao, Wu, Yu, and
  Wang]{liu2026can}
J.~Liu, F.~Gao, B.~Wei, X.~Chen, Q.~Liao, Y.~Wu, C.~Yu, and Y.~Wang.
\newblock What can rl bring to vla generalization? an empirical study.
\newblock \emph{Advances in Neural Information Processing Systems},
  38:\penalty0 97121--97151, 2026.

\bibitem[Kumar et~al.(2021)Kumar, Hong, Singh, and Levine]{kumar2021should}
A.~Kumar, J.~Hong, A.~Singh, and S.~Levine.
\newblock Should i run offline reinforcement learning or behavioral cloning?
\newblock In \emph{International conference on learning representations}, 2021.

\bibitem[Williams et~al.(2017)Williams, Aldrich, and
  Theodorou]{williams2017model}
G.~Williams, A.~Aldrich, and E.~A. Theodorou.
\newblock Model predictive path integral control: From theory to parallel
  computation.
\newblock \emph{Journal of Guidance, Control, and Dynamics}, 40\penalty0
  (2):\penalty0 344--357, 2017.

\bibitem[Hansen et~al.(2022)Hansen, Wang, and Su]{hansen2022temporal}
N.~Hansen, X.~Wang, and H.~Su.
\newblock Temporal difference learning for model predictive control.
\newblock 2022.

\bibitem[Hansen et~al.(2024)Hansen, Su, and Wang]{hansen2023td}
N.~Hansen, H.~Su, and X.~Wang.
\newblock Td-mpc2: Scalable, robust world models for continuous control.
\newblock 2024.

\bibitem[Schrittwieser et~al.(2020)Schrittwieser, Antonoglou, Hubert, Simonyan,
  Sifre, Schmitt, Guez, Lockhart, Hassabis, Graepel,
  et~al.]{schrittwieser2020mastering}
J.~Schrittwieser, I.~Antonoglou, T.~Hubert, K.~Simonyan, L.~Sifre, S.~Schmitt,
  A.~Guez, E.~Lockhart, D.~Hassabis, T.~Graepel, et~al.
\newblock Mastering atari, go, chess and shogi by planning with a learned
  model.
\newblock \emph{Nature}, 588\penalty0 (7839):\penalty0 604--609, 2020.

\bibitem[Oquab et~al.(2023)Oquab, Darcet, Moutakanni, Vo, Szafraniec, Khalidov,
  Fernandez, Haziza, Massa, El-Nouby, et~al.]{oquab2023dinov2}
M.~Oquab, T.~Darcet, T.~Moutakanni, H.~Vo, M.~Szafraniec, V.~Khalidov,
  P.~Fernandez, D.~Haziza, F.~Massa, A.~El-Nouby, et~al.
\newblock Dinov2: Learning robust visual features without supervision.
\newblock \emph{arXiv preprint arXiv:2304.07193}, 2023.

\bibitem[Raffel et~al.(2020)Raffel, Shazeer, Roberts, Lee, Narang, Matena,
  Zhou, Li, and Liu]{raffel2020exploring}
C.~Raffel, N.~Shazeer, A.~Roberts, K.~Lee, S.~Narang, M.~Matena, Y.~Zhou,
  W.~Li, and P.~J. Liu.
\newblock Exploring the limits of transfer learning with a unified text-to-text
  transformer.
\newblock \emph{Journal of machine learning research}, 21\penalty0
  (140):\penalty0 1--67, 2020.

\bibitem[Brohan et~al.(2022)Brohan, Brown, Carbajal, Chebotar, Dabis, Finn,
  Gopalakrishnan, Hausman, Herzog, Hsu, et~al.]{brohan2022rt}
A.~Brohan, N.~Brown, J.~Carbajal, Y.~Chebotar, J.~Dabis, C.~Finn,
  K.~Gopalakrishnan, K.~Hausman, A.~Herzog, J.~Hsu, et~al.
\newblock Rt-1: Robotics transformer for real-world control at scale.
\newblock \emph{arXiv preprint arXiv:2212.06817}, 2022.

\bibitem[Padalkar et~al.(2023)Padalkar, Pooley, Jain, Bewley, Herzog, Irpan,
  Khazatsky, Rai, Singh, Brohan, et~al.]{padalkar2023open}
A.~Padalkar, A.~Pooley, A.~Jain, A.~Bewley, A.~Herzog, A.~Irpan, A.~Khazatsky,
  A.~Rai, A.~Singh, A.~Brohan, et~al.
\newblock Open x-embodiment: Robotic learning datasets and rt-x models.
\newblock \emph{arXiv preprint arXiv:2310.08864}, 2023.

\bibitem[Bousmalis et~al.(2023)Bousmalis, Vezzani, Rao, Devin, Lee, Bauza,
  Davchev, Zhou, Gupta, Raju, et~al.]{bousmalis2023robocat}
K.~Bousmalis, G.~Vezzani, D.~Rao, C.~Devin, A.~X. Lee, M.~Bauza, T.~Davchev,
  Y.~Zhou, A.~Gupta, A.~Raju, et~al.
\newblock Robocat: A self-improving foundation agent for robotic manipulation.
\newblock \emph{arXiv preprint arXiv:2306.11706}, 2023.

\bibitem[Laskey et~al.(2017)Laskey, Lee, Fox, Dragan, and
  Goldberg]{laskey2017dart}
M.~Laskey, J.~Lee, R.~Fox, A.~Dragan, and K.~Goldberg.
\newblock Dart: Noise injection for robust imitation learning.
\newblock In \emph{Conference on robot learning}, pages 143--156. PMLR, 2017.

\bibitem[Mnih et~al.(2013)Mnih, Kavukcuoglu, Silver, Graves, Antonoglou,
  Wierstra, and Riedmiller]{mnih2013playing}
V.~Mnih, K.~Kavukcuoglu, D.~Silver, A.~Graves, I.~Antonoglou, D.~Wierstra, and
  M.~Riedmiller.
\newblock Playing atari with deep reinforcement learning.
\newblock \emph{arXiv preprint arXiv:1312.5602}, 2013.

\bibitem[Haarnoja et~al.(2018)Haarnoja, Zhou, Hartikainen, Tucker, Ha, Tan,
  Kumar, Zhu, Gupta, Abbeel, et~al.]{haarnoja2018soft}
T.~Haarnoja, A.~Zhou, K.~Hartikainen, G.~Tucker, S.~Ha, J.~Tan, V.~Kumar,
  H.~Zhu, A.~Gupta, P.~Abbeel, et~al.
\newblock Soft actor-critic algorithms and applications.
\newblock \emph{arXiv preprint arXiv:1812.05905}, 2018.

\bibitem[Lillicrap et~al.(2015)Lillicrap, Hunt, Pritzel, Heess, Erez, Tassa,
  Silver, and Wierstra]{lillicrap2015continuous}
T.~P. Lillicrap, J.~J. Hunt, A.~Pritzel, N.~Heess, T.~Erez, Y.~Tassa,
  D.~Silver, and D.~Wierstra.
\newblock Continuous control with deep reinforcement learning.
\newblock \emph{arXiv preprint arXiv:1509.02971}, 2015.

\bibitem[Nakamoto et~al.(2024)Nakamoto, Mees, Kumar, and
  Levine]{nakamoto2024steering}
M.~Nakamoto, O.~Mees, A.~Kumar, and S.~Levine.
\newblock Steering your generalists: Improving robotic foundation models via
  value guidance.
\newblock \emph{arXiv preprint arXiv:2410.13816}, 2024.

\bibitem[Bhateja et~al.(2023)Bhateja, Guo, Ghosh, Singh, Tomar, Vuong,
  Chebotar, Levine, and Kumar]{bhateja2023robotic}
C.~Bhateja, D.~Guo, D.~Ghosh, A.~Singh, M.~Tomar, Q.~Vuong, Y.~Chebotar,
  S.~Levine, and A.~Kumar.
\newblock Robotic offline rl from internet videos via value-function
  pre-training.
\newblock \emph{arXiv preprint arXiv:2309.13041}, 2023.

\bibitem[Ma et~al.(2022)Ma, Sodhani, Jayaraman, Bastani, Kumar, and
  Zhang]{ma2022vip}
Y.~J. Ma, S.~Sodhani, D.~Jayaraman, O.~Bastani, V.~Kumar, and A.~Zhang.
\newblock Vip: Towards universal visual reward and representation via
  value-implicit pre-training.
\newblock \emph{arXiv preprint arXiv:2210.00030}, 2022.

\bibitem[Ma et~al.(2023)Ma, Kumar, Zhang, Bastani, and Jayaraman]{ma2023liv}
Y.~J. Ma, V.~Kumar, A.~Zhang, O.~Bastani, and D.~Jayaraman.
\newblock Liv: Language-image representations and rewards for robotic control.
\newblock In \emph{International Conference on Machine Learning}, pages
  23301--23320. PMLR, 2023.

\bibitem[Wagenmaker et~al.(2025)Wagenmaker, Nakamoto, Zhang, Park, Yagoub,
  Nagabandi, Gupta, and Levine]{wagenmaker2025steering}
A.~Wagenmaker, M.~Nakamoto, Y.~Zhang, S.~Park, W.~Yagoub, A.~Nagabandi,
  A.~Gupta, and S.~Levine.
\newblock Steering your diffusion policy with latent space reinforcement
  learning.
\newblock \emph{arXiv preprint arXiv:2506.15799}, 2025.

\bibitem[Xiao et~al.(2025)Xiao, Lin, Peng, Xue, He, Xie, Hu, Wu, Luo, Fan,
  et~al.]{xiao2025self}
W.~Xiao, H.~Lin, A.~Peng, H.~Xue, T.~He, Y.~Xie, F.~Hu, J.~Wu, Z.~Luo, L.~Fan,
  et~al.
\newblock Self-improving vision-language-action models with data generation via
  residual rl.
\newblock \emph{arXiv preprint arXiv:2511.00091}, 2025.

\bibitem[Kumar et~al.(2022)Kumar, Singh, Ebert, Nakamoto, Yang, Finn, and
  Levine]{kumar2022pre}
A.~Kumar, A.~Singh, F.~Ebert, M.~Nakamoto, Y.~Yang, C.~Finn, and S.~Levine.
\newblock Pre-training for robots: Offline rl enables learning new tasks from a
  handful of trials.
\newblock \emph{arXiv preprint arXiv:2210.05178}, 2022.

\bibitem[Yang et~al.(2024)Yang, Mark, Vu, Sharma, Bohg, and
  Finn]{yang2024robot}
J.~Yang, M.~S. Mark, B.~Vu, A.~Sharma, J.~Bohg, and C.~Finn.
\newblock Robot fine-tuning made easy: Pre-training rewards and policies for
  autonomous real-world reinforcement learning.
\newblock In \emph{2024 IEEE International Conference on Robotics and
  Automation (ICRA)}, pages 4804--4811. IEEE, 2024.

\bibitem[Hu et~al.(2025)Hu, Wu, Enock, Li, Kadakia, Erickson, and
  Kumar]{hu2025rac}
Z.~Hu, R.~Wu, N.~Enock, J.~Li, R.~Kadakia, Z.~Erickson, and A.~Kumar.
\newblock Rac: Robot learning for long-horizon tasks by scaling recovery and
  correction.
\newblock \emph{arXiv preprint arXiv:2509.07953}, 2025.

\bibitem[Guo et~al.(2026)Guo, Lee, Shi, Chen, Liang, and Finn]{guo2026vlaw}
Y.~Guo, T.~Lee, L.~X. Shi, J.~Chen, P.~Liang, and C.~Finn.
\newblock Vlaw: Iterative co-improvement of vision-language-action policy and
  world model.
\newblock \emph{arXiv preprint arXiv:2602.12063}, 2026.

\bibitem[Liu et~al.(2026)Liu, Tan, Zhu, Li, Li, Yang, and Shen]{liu2026self}
C.~Liu, W.~Tan, L.~Zhu, F.~Li, J.~Li, G.~Yang, and H.~T. Shen.
\newblock Self-correcting vla: Online action refinement via sparse world
  imagination.
\newblock \emph{arXiv preprint arXiv:2602.21633}, 2026.

\bibitem[Jia et~al.(2026)Jia, Yuan, Shi, Guizilini, Mao, and
  Wang]{jia2026dreamplan}
E.~Y.-T. Jia, W.~Yuan, T.~Shi, V.~Guizilini, J.~Mao, and Y.~Wang.
\newblock Dreamplan: Efficient reinforcement fine-tuning of vision-language
  planners via video world models.
\newblock \emph{arXiv preprint arXiv:2603.16860}, 2026.

\bibitem[Hu et~al.(2023)Hu, Mirchandani, and Sadigh]{hu2023ibrl}
H.~Hu, S.~Mirchandani, and D.~Sadigh.
\newblock Imitation bootstrapped reinforcement learning.
\newblock \emph{arXiv preprint arXiv:2311.02198}, 2023.

\bibitem[Ren et~al.(2024)Ren, Lidard, Ankile, Simeonov, Agrawal, Majumdar,
  Burchfiel, Dai, and Simchowitz]{ren2024dppo}
A.~Z. Ren, J.~Lidard, L.~L. Ankile, A.~Simeonov, P.~Agrawal, A.~Majumdar,
  B.~Burchfiel, H.~Dai, and M.~Simchowitz.
\newblock Diffusion policy policy optimization.
\newblock \emph{arXiv preprint arXiv:2409.00588}, 2024.

\bibitem[Dodeja et~al.(2026)]{dodeja2026bcq}
L.~Dodeja et~al.
\newblock When life gives you {BC}, make {Q}-functions: Extracting {Q}-values
  from behavior cloning for on-robot reinforcement learning.
\newblock \emph{arXiv preprint arXiv:2605.05172}, 2026.

\bibitem[Farebrother et~al.(2024)Farebrother, Orbay, Vuong, Ta{\"\i}ga,
  Chebotar, Xiao, Irpan, Levine, Castro, Faust, et~al.]{farebrother2024stop}
J.~Farebrother, J.~Orbay, Q.~Vuong, A.~A. Ta{\"\i}ga, Y.~Chebotar, T.~Xiao,
  A.~Irpan, S.~Levine, P.~S. Castro, A.~Faust, et~al.
\newblock Stop regressing: Training value functions via classification for
  scalable deep rl.
\newblock \emph{arXiv preprint arXiv:2403.03950}, 2024.

\bibitem[Li et~al.(2026)Li, Zhou, and Levine]{li2026reinforcement}
Q.~Li, Z.~P. Zhou, and S.~Levine.
\newblock Reinforcement learning with action chunking.
\newblock \emph{Advances in Neural Information Processing Systems},
  38:\penalty0 55518--55553, 2026.

\bibitem[Liu et~al.(2023)Liu, Zhu, Gao, Feng, Liu, Zhu, and
  Stone]{liu2023libero}
B.~Liu, Y.~Zhu, C.~Gao, Y.~Feng, Q.~Liu, Y.~Zhu, and P.~Stone.
\newblock Libero: Benchmarking knowledge transfer for lifelong robot learning.
\newblock \emph{Advances in Neural Information Processing Systems},
  36:\penalty0 44776--44791, 2023.

\bibitem[Mu et~al.(2025)Mu, Chen, Chen, Peng, Lan, Gao, Liang, Yu, Zou, Xu,
  et~al.]{mu2025robotwin}
Y.~Mu, T.~Chen, Z.~Chen, S.~Peng, Z.~Lan, Z.~Gao, Z.~Liang, Q.~Yu, Y.~Zou,
  M.~Xu, et~al.
\newblock Robotwin: Dual-arm robot benchmark with generative digital twins.
\newblock In \emph{Proceedings of the computer vision and pattern recognition
  conference}, pages 27649--27660, 2025.

\end{thebibliography}

\clearpage
\appendix

\section{Q-function architecture}
\label{app:arch}

\begin{figure}[H]
    \centering
    \begin{tikzpicture}[
        >={Stealth[length=3.5pt,width=3.5pt]},
        every node/.style={font=\footnotesize},
        input/.style={rectangle, rounded corners=2pt, draw=gray!60, fill=gray!6,
                      minimum height=0.55cm, minimum width=1.45cm, align=center,
                      line width=0.4pt, inner sep=4pt},
        qparam/.style={rectangle, rounded corners=2pt, draw=blue!70!black, fill=blue!10,
                       minimum height=0.85cm, minimum width=1.95cm, align=center,
                       line width=0.6pt, inner sep=6pt},
        output/.style={rectangle, rounded corners=2pt, draw=orange!70!black, fill=orange!10,
                       minimum height=0.55cm, minimum width=1.05cm, align=center,
                       line width=0.5pt, inner sep=4pt},
        arrow/.style={->, line width=0.45pt},
        node distance=0.5cm and 1.1cm,
    ]
        \node[input] (img)  at (0,  1.45) {images $\vo_t$};
        \node[input] (lang) at (0,  0.00) {language $\vell$};
        \node[input] (act)  at (0, -1.75) {action chunk\\$\va_{t:t+H}$};

        \node[qparam] (dino) at (2.6,  1.45) {DinoV2};
        \node[qparam] (t5)   at (2.6,  0.00) {T5};

        \node[qparam, minimum width=2.3cm, minimum height=1.8cm, align=center, inner sep=6pt]
            (dec) at (6.8, -0.15) {Transformer\\decoder\\\scriptsize (cross-attn)};

        \node[qparam, minimum width=1.95cm] (hg) at (9.6, -0.15)
            {HL-Gauss head\\\scriptsize ($B$ bin logits)};

        \node[output] (q) at (11.6, -0.15) {$Q_\phi$};

        \draw[arrow] (img.east)  -- (dino.west);
        \draw[arrow] (lang.east) -- (t5.west);

        \coordinate (kv) at (5.15, 0.75);
        \draw[line width=0.45pt] (dino.east) -|
            node[pos=0.25, above, font=\scriptsize, yshift=-1pt] {$\vz^{\text{vis}}_t$} (kv);
        \draw[line width=0.45pt] (t5.east)   -|
            node[pos=0.25, above, font=\scriptsize, yshift=-1pt] {$\vz^{\text{txt}}$} (kv);
        \draw[arrow] (kv) |- (dec.west);

        \draw[arrow] (act.east) -| (dec.south)
            node[pos=0.78, right, font=\scriptsize, xshift=-1pt, yshift=-3pt] {query tokens};

        \draw[arrow] (dec.east) -- (hg.west);
        \draw[arrow] (hg.east)  -- node[above, font=\scriptsize, yshift=-1pt] {$\mathbb{E}$} (q.west);
    \end{tikzpicture}
    \caption{\textbf{Q-function architecture.} The Q-function $Q_\phi$ has its own DinoV2 visual encoder and T5 language encoder (parameter-disjoint from the BC policy). The transformer decoder cross-attends to visual tokens $\vz^{\text{vis}}_t$ and language tokens $\vz^{\text{txt}}$ (keys/values) and takes the candidate action chunk $\va_{t:t+H}$ as a sequence of query tokens. The HL-Gauss head outputs $B$ bin logits over discounted returns and the scalar $Q_\phi$ is the expectation of the softmaxed bin distribution.}
    \label{fig:arch}
\end{figure}
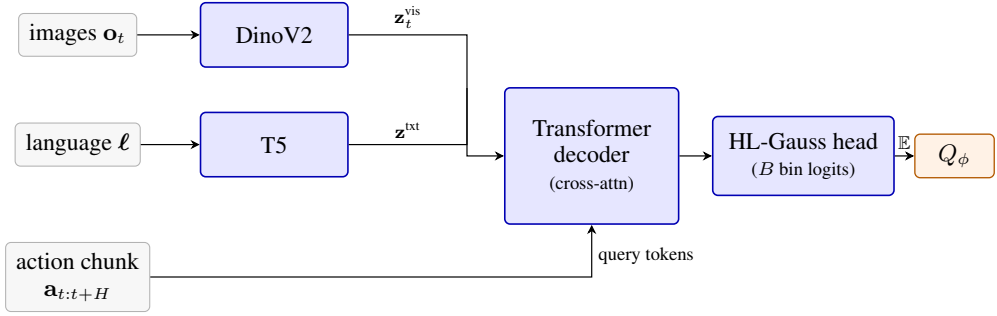

\section{Implementation details}
\label{app:impl}

\textbf{Hyperparameters.} We use $H = 32$ action-chunk length, inherited from FastWAM's action
horizon~\citep{yuan2026fast}, $N = 64$ candidate chunks per planning step on
LIBERO and $N = 32$ on RoboTwin, Q-weighting temperature $\lambda = 1$, and
discount $\gamma = 0.99$. The planner executes the first $10$ steps of each
chunk on LIBERO and the first $24$ on RoboTwin before replanning
(App.~\ref{app:latency}). When sampling action chunks from FastWAM during \qplan{}, we use $3$ flow-matching denoising steps (rather than the default $10$), which lowers latency and preserves the sample diversity that a Q-weighted average benefits from. The Q-function output uses $B = 101$ HL-Gauss bins on $[v_{\min}, v_{\max}] = [0, 1]$ with kernel width matched to bin spacing. Optimisation uses AdamW with learning rate $3{\times}10^{-4}$ and an EMA target rate of $\eta = 0.005$. Q-training runs on $4 \times$ H200 GPUs for $12$k gradient steps on LIBERO and $45$k on RoboTwin before evaluation. One self-improvement iteration consists of $M = 100$ collected episodes per task (20 per task on the real robot) followed by $S = 200$ Q-only gradient steps on the combined offline + online buffer, with each minibatch drawn half from the original demonstrations and half from the accumulated online rollouts; the BC policy is frozen for all $S$ steps and across all iterations. LIBERO episodes are capped at $540$ environment steps; RoboTwin episodes use the benchmark default.

\textbf{Model size.} The Q-function totals approximately $1$B parameters (DinoV2 visual encoder $\sim 300$M, T5 language encoder $\sim 220$M, transformer decoder $\sim 500$M). This is small relative to the multi-billion-parameter FastWAM policy, and accounts for why one self-improvement iteration ($S{=}200$ Q-only gradient steps) is dramatically cheaper than a full-policy gradient step.

\textbf{Training stability.} Two ingredients keep Q-training stable under sparse terminal rewards: HL-Gauss categorical regression removes the scale sensitivity of scalar MSE regression to sparse, bimodal returns, and Q-chunking shrinks the effective bootstrapping horizon by a factor of $H$, mitigating long-horizon variance blow-up.

\section{Bootstrap target: one-step-shifted vs planner-selected}
\label{app:bootstrap}

The Bellman loss in Eq.~\ref{eq:q-target} uses the next buffer chunk $\va_{t+H:t+2H}$ as the bootstrap action rather than an action selected by our current planner. Two considerations justify this choice. First, it removes a full planner call from every Bellman update, which makes training an order of magnitude cheaper than an explicit Q-learning update~\citep{li2026reinforcement}. Second, as self-improvement proceeds the buffer fills with on-policy rollouts collected by the Q-guided planner, so the shifted target effectively bootstraps from planner-selected actions and the value/policy gap closes without an explicit correction. We tried a planner-selected bootstrap variant and observed no gain; we keep the shifted target for simplicity and training speed.

\section{Temporal-smoothed MPPI (baseline in Sec.~\ref{sec:exp-planning})}
\label{app:temporal}

Two of the rows in Table~\ref{tab:planning} use variants of Model Predictive Path Integral (MPPI) control~\citep{williams2017model} as the action source that the Q-function scores. We describe them here for completeness.

At each planning step $t$ we run $T$ iterations of MPPI. At iteration $i$, the proposal distribution over action chunks has mean $\bar{\va}^{(i)}_{t:t+H} \in \R^{H \times d_a}$, initialised at iteration $i{=}0$ to the BC sample $\bar{\va}^{(0)}_{t:t+H} = \pi^{\text{BC}}(\vo_t, \vell)$ (the BC \emph{warm-start}), and is perturbed by additive Gaussian noise:
\begin{equation}
\label{eq:mppi-sample}
\begin{aligned}
\va^{(n,i)}_{t:t+H} &= \bar{\va}^{(i)}_{t:t+H} + G_\sigma \ast_k \boldsymbol{\xi}^{(n)}, \\
\boldsymbol{\xi}^{(n)} &\in \R^{H \times d_a}, \;\; \boldsymbol{\xi}^{(n)}_{k,j} \sim \mathcal{N}(0,1), \;\; n=1,\ldots,N,
\end{aligned}
\end{equation}
where $G_\sigma$ is a $1$-D Gaussian kernel of standard deviation $\sigma$ convolved along the chunk axis $k$. When $\sigma \to 0$ the kernel is the identity and we recover the ``vanilla MPPI'' row of Table~\ref{tab:planning}: independent Gaussian perturbations at every timestep of the chunk, which produce jagged trajectories whose per-step actions can jump far from the BC's mean. Setting $\sigma > 0$ (we use $\sigma = 2$ over $H = 32$) correlates the noise across neighbouring timesteps so perturbed chunks stay temporally coherent and lie closer to the BC's own trajectory manifold, which is what recovers the $+3$pp gain in Table~\ref{tab:planning}.

MPPI weights are then computed from the Q-scores with temperature $\lambda$ and used to update the proposal mean:
\begin{equation}
\label{eq:mppi-weight}
w^{(n,i)} \propto \exp\!\big(Q_\phi(\vo_t, \vell, \va^{(n,i)}_{t:t+H}) / \lambda\big),
\qquad
\bar{\va}^{(i+1)}_{t:t+H} = \sum_{n=1}^N w^{(n,i)} \, \va^{(n,i)}_{t:t+H} .
\end{equation}
After $T$ iterations, the converged mean $\bar{\va}^{(T)}_{t:t+H}$ is executed in the environment. Our current \qplan{} planner (Sec.~\ref{sec:method-plan}) is the limit $T{=}1$ with $\sigma$ made irrelevant by drawing the candidates directly from a $3$-step flow-matching pass of the BC head, whose native multi-modality plays the role that temporally-smoothed noise plays here.

\section{Planning-step latency profile}
\label{app:latency}

We profile a full \qplan{} planning step (one action chunk) on a single NVIDIA L40S GPU, batch $1$, median $+$ p95 over $60$ timed steps after $10$ warm-up steps, with \texttt{torch.cuda.synchronize} around every timed region. The strict setting uses \texttt{lerobot-eval} parity (deterministic algorithms, TF32 off, \texttt{cudnn.benchmark} off, BC in bf16, Q in fp32) which reproduces the reported success rates; the standard-inference setting turns TF32 on, enables \texttt{cudnn.benchmark}, and drops deterministic kernels (a deployment mode). The replanning cadence is the deadline a planning step must beat: the
planner emits an $H{=}32$-step chunk, of which the first $10$ (LIBERO,
$30$\,Hz) or $24$ (RoboTwin, $25$\,Hz) steps execute before replanning,
giving budgets of $333$ ms and $960$ ms.

\begin{table}[H]
\centering
\scriptsize
\setlength{\tabcolsep}{3pt}
\renewcommand{\arraystretch}{1.1}
\caption{\textbf{Planning-step latency (ms) on an NVIDIA L40S.} Median over $60$ timed steps; parenthesised numbers are p95. Percentages express the fraction of the replan budget (LIBERO $333$ ms, RoboTwin $960$ ms). Strict = eval parity settings. Rows marked $\blacktriangleleft$ are the deployed configurations. The final row is an earlier iterative sample-based variant we prototyped, retained here for comparison.}
\label{tab:latency}
\begin{tabular}{@{}lccccc@{}}
\toprule
Configuration & Q-evals & LIBERO strict & \% & RoboTwin strict & \% \\
\midrule
BC only, $3$ denoise steps & $0$ & $273$ ($275$) & $82$ & $273$ ($275$) & $28$ \\
BC only, $10$ steps (eval baseline)      & $0$ & $646$ ($647$) & $194$ & $640$ ($643$) & $67$ \\
Q-Planning, $N{=}16$, $3$ steps          & $16$ & $337$ ($338$) & $101$ & $352$ ($354$) & $37$ \\
Q-Planning, $N{=}32$, $3$ steps $\blacktriangleleft$ RoboTwin & $32$ & $371$ ($372$) & $111$ & $400$ ($402$) & $42$ \\
Q-Planning, $N{=}64$, $3$ steps $\blacktriangleleft$ LIBERO  & $64$ & $640$ ($641$) & $192$ & $716$ ($725$) & $75$ \\
Earlier iterative sample-based variant ($3$ iters, $N{=}64$) & $192$ & $1114$ ($1117$) & $334$ & $1276$ ($1282$) & $133$ \\
\bottomrule
\end{tabular}
\end{table}

\textbf{Findings.} (i)~Under strict eval-parity settings, RoboTwin is comfortably real-time: the deployed $N{=}32$ planner uses only $42\%$ of the $960$ ms budget, and $3.2\times$ faster than the earlier iterative variant which would have missed the budget at $133\%$. (ii)~LIBERO is latency-neutral relative to its own frozen-BC baseline: the deployed $N{=}64$ configuration takes $640$ ms per step, essentially matching the $646$ ms cost of a single $10$-step BC draw the baseline itself pays. Lowering to $N{=}16$ fits the $333$ ms budget with comparable success. (iii)~Encoders (DinoV2 $+$ T5) run once per planning step at $23$-$27$ ms; only the $\sim 500$M Q decoder scales with $N$ (roughly $2$-$3$ ms per candidate), which is why doubling $N$ does not double the total step cost.

\section{Real-robot setup}
\label{app:real}

\textbf{Platform and control.} Two $6$-DoF YAM arms, table-mounted and
forward-facing (Fig.~\ref{fig:si-real}). Observations are RGB images from
three cameras, one overhead and one on each wrist, shared by the BC policy and
the Q-function. Actions are joint-space commands for both arms; the BC emits
$H{=}30$ action chunks ($1$ s at $30$ Hz) and the planner executes the full
chunk open-loop before replanning, giving a $1$ s replan budget. Policy inference runs on a single RTX 5090.

\textbf{Data and evaluation protocol.} Each task starts from $100$
demonstrations. Initial conditions are randomised within the demonstration
distribution: cup positions vary across episodes, and for insert-wallet the
card's starting slot varies while the wallet stays fixed. Each
self-improvement iteration collects $20$ episodes per task, and the
per-iteration success rates in Fig.~\ref{fig:si-real} are measured on these
collection episodes. The robot is never teleoperated during rollouts; human
input during the loop is limited to scene resets between episodes and the
per-episode success label that supplies the terminal reward
(Sec.~\ref{sec:method-setup}).

\textbf{SFT baseline.} The SFT comparison in Fig.~\ref{fig:si-real} fully
fine-tunes the policy on the successful rollouts collected under the
same episode budget, discarding failures; \qplan{} instead absorbs the full
rollout set into Q-only updates with the hyperparameters of
App.~\ref{app:impl}.

\end{document}